\documentclass{article}
\usepackage{ijcai26}
\usepackage{times}
\usepackage{soul}
\usepackage{url}
\usepackage[hidelinks]{hyperref}
\usepackage[utf8]{inputenc}
\usepackage[small]{caption}
\usepackage{graphicx}
\usepackage{amsmath}
\usepackage{amsthm}
\usepackage{booktabs}
\usepackage{algorithm}
\usepackage{algorithmic}
\usepackage[switch]{lineno}
\usepackage{amsfonts}
\usepackage{multirow}
\usepackage[table]{xcolor}

\usepackage{tikz}
\usepackage[edges]{forest}
\definecolor{hidden-draw}{RGB}{0,0,0}
\definecolor{hidden-pink}{rgb}{0.98, 0.94, 0.75}
\definecolor{level0}{rgb}{0.98, 0.92, 0.84}
\definecolor{level1}{rgb}{1.0, 0.71, 0.76}
\definecolor{level2}{rgb}{0.67, 0.88, 0.69}
\definecolor{level3}{rgb}{0.56, 0.87, 0.98}
\definecolor{level4}{rgb}{0.98, 0.92, 0.84}
\definecolor{darkred}{RGB}{200, 0, 0}
\definecolor{darkgreen}{RGB}{0, 100, 0}
\usetikzlibrary{trees, positioning, shapes.multipart}

\usepackage[table]{xcolor} 
\usepackage[most]{tcolorbox}
\newtcolorbox{mybox}{
    colback=white,
    colframe=black!75!white,
    boxrule=1pt,
    arc=4pt,
    left=6pt, right=6pt, top=6pt, bottom=6pt,
    breakable
}

\nolinenumbers

\title{From Time Series Analysis to Question Answering: A Survey in the LLM Era}

\author{
Wei Li$^1$\and
Zhe Xie$^2$\and
Yuxuan Liang$^3$\and
Xinli Hao$^1$\and
Yunyao Cheng$^4$\and
Dan Pei$^2$\and
Xiaofeng Meng$^1$\footnote{Corresponding Author}\\
\affiliations
$^1$Renmin University of China\\
$^2$Tsinghua University\\
$^3$Hong Kong University of Science and Technology (Guangzhou)\\
$^4$Aalborg University\\
\emails
\{leeway, xinli\_hao, xfmeng\}@ruc.edu.cn,
\{xiez22, peidan\}@tsinghua.edu.cn,\\
yuxliang@outlook.com,yunyaoc@cs.aau.dk
}

\begin{document}

\maketitle

\begin{abstract}
Recently, Large Language Models (LLMs) have introduced a novel paradigm in Time Series Analysis (TSA), leveraging strong language capabilities to support tasks such as forecasting and anomaly detection. However, these analysis tasks cannot adequately cover temporal language tasks, such as interpretation and captioning. A fundamental gap remains between TSA and LLMs: LLMs are pre-trained to optimize natural language relevance for question answering rather than objectives specialized for TSA. To bridge this gap, TSA is evolving toward Time Series Question Answering (TSQA), shifting from expert-driven and task-specific analysis to user-driven and task-unified question answering. TSQA depends on flexible exploration rather than predefined TSA pipelines. In this survey, we first propose a taxonomy that reflects the evolution from TSA to TSQA, driven by a shift from external to internal alignment. We then organize existing literature into three alignment paradigms: Injective Alignment, Bridging Alignment, and Internal Alignment, and provide practical guidance for flexible, economical, and generalizable selection of alignment paradigms. We finally analyze datasets across domains and characteristics, identify challenges, and highlight future research directions.

\end{abstract}

\section{Introduction}
Time series are sequential data characterized by temporal dependencies and pattern semantics~\cite{Survey_FMTSA}. Time Series Analysis (TSA) has been widely applied across multiple domains, including healthcare~\cite{MedTsLLM}, finance~\cite{SCRL-LG}, IoT~\cite{ECG-LLM}, weather~\cite{YunyaoCheng_1}, and electricity~\cite{LLM-TPF}. Within these domains, TSA typically involves tasks such as forecasting~\cite{Time-LLM}, anomaly detection~\cite{AnomalyLLM}, classification~\cite{TableTime}, and imputation~\cite{Survey_Imputation}. Recently, Large Language Models (LLMs) have introduced a novel paradigm in TSA, leveraging strong natural language capabilities~\cite{Survey_TSLLM}. 

\begin{figure}[!t]
\centering
\includegraphics[width=\linewidth]{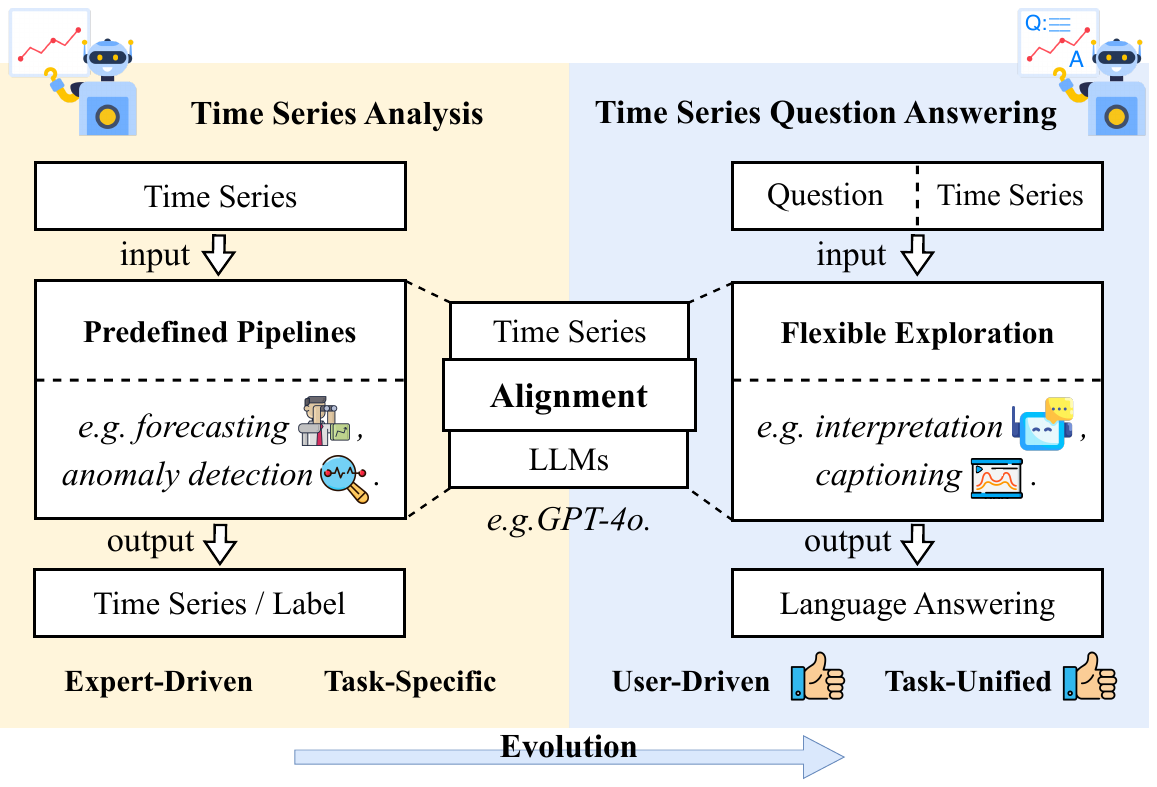}
\caption{Comparison between Time Series Analysis and Time Series Question Answering under alignment of time series with LLMs.}
\label{fig:motivation}
\vspace{-0.2em}
\end{figure}

However, these analysis tasks cannot adequately cover temporal language tasks. For example, in database AIOps, Time Series Question Answering (TSQA) provides comprehensive interpretations of performance metrics~\cite{ChatTS}. In time series captioning tasks, TSQA summarizes temporal patterns, structures, and events~\cite{TSLM}. A fundamental gap remains between TSA and LLMs: LLMs are pre-trained to optimize natural language relevance for question answering rather than objectives specialized for TSA.
As shown in Figure~\ref{fig:motivation}, TSA is evolving toward TSQA, shifting from expert-driven and task-specific analysis to user-driven and task-unified question answering.
TSA relies on predefined pipelines built on task-specific time series modeling. In contrast, TSQA depends on flexible exploration based on task-unified temporal language processing, including in-depth interpretation and captioning~\cite{Survey_AgenticTSLLM}.
After aligning time series with LLMs (e.g., GPT-4o~\cite{gpt-4o}), user-driven TSQA inputs time series and questions and outputs natural language answers, which are more intuitive than the numerical time series or categorical labels generated by expert-driven TSA~\cite{Survey_TSLLM}.
The fundamental gap highlights alignment between time series and LLMs as a central challenge, motivating us to propose alignment paradigms that cover the evolution from TSA to TSQA.

\begin{figure*}[!t]
\centering
\includegraphics[width=\linewidth]{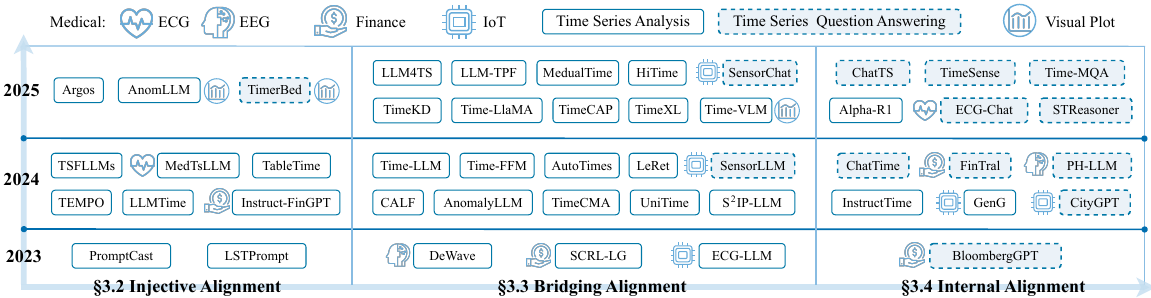}
\caption{Taxonomy of relevant literature across three alignment paradigms. The icon to the left of each method indicates its domain, while the absence of this icon denotes a general domain. The icon on the right denotes the use of time series visual plots. The background color of each method represents task targets: lighter shades denote time series analysis (TSA), and darker shades denote time series question answering (TSQA). The TSA literature dominates external Injective and Bridging Alignment, whereas the TSQA literature dominates Internal Alignment, reflecting \textbf{the evolution from TSA to TSQA driven by a shift from external to internal alignment}.}
\label{fig:taxonomy}
\vspace{-0.2em}
\end{figure*}

The perspectives of representative survey literature are summarized in Table~\ref{tab:survey}. Instead of considering TSA and TSQA in isolation, our survey adopts the evolution perspective from TSA to TSQA and provides an up-to-date overview. Compared with existing surveys, this evolution perspective integrates previously isolated TSA and TSQA research into a unified taxonomy of alignment paradigms in the LLM era.

As shown in Figure~\ref{fig:taxonomy}, the taxonomy consists of three alignment paradigms:  
(1) \textbf{Injective Alignment} injects numerical time series into textual prompts without temporal modification, using frozen LLMs. Temporal modification refers to optimizations that adapt time series to LLMs.
(2) \textbf{Bridging Alignment} establishes mappings between numerical time series and prompts, introducing temporal modification while still employing frozen LLMs.  
(3) \textbf{Internal Alignment} leverages multiple time series representations, including numerical values, visual plots, and structured tables, combining temporal modification with internal LLM components.

The primary contributions are summarized as follows.
\begin{itemize}
\item We distinguish and define TSA and TSQA (Section~\ref{sec:pre}). 
\item We propose a taxonomy of three alignment paradigms, provide practical guidance for paradigm selection, and systematically categorize the relevant literature and maintain an up-to-date GitHub repository\footnote{\url{https://github.com/Leeway-95/TSA-TSQA-with-LLMs}} (Section~\ref{sec:taxonomy}). 
\item We analyze datasets from the perspectives of application domains and data characteristics (Section~\ref{sec:dataset}).
\item We discuss challenges and future directions (Section~\ref{sec:future}).
\end{itemize}

\begin{table}[ht]
\centering
\begin{tabular}{@{}c|c@{}}
\toprule
\textbf{Survey Literature} & \textbf{Perspective} \\ 
\midrule
Our Survey & TSA\&TSQA Evolution \\ \midrule
\cite{Survey_CMA} (IJCAI'25) & Cross-modality TSA\\  \midrule
\cite{Survey_MTSA} (KDD'25) & Multi-modality TSA\\  \midrule
\cite{Survey_TSLLM} (IJCAI'24) & Pipeline-based TSA\\  \midrule
\cite{Survey_FMTSA} (KDD'24) & Foundation Model \\
\bottomrule
\end{tabular}
\caption{Comparison between our perspective and existing surveys.}
\label{tab:survey}
\vspace{-0.3em}
\end{table}

\section{Preliminaries}
\label{sec:pre}
\subsection{Definitions}
A time series is represented as a numerical sequence with temporal order $\mathbf{X}_\text{Number} = \langle\mathbf{x}_{1}, \ldots, \mathbf{x}_{S}\rangle \in \mathbb{R}^{S \times N}$, where $S$ denotes the sequence length and $N$ is the number of variables. Each observation $\mathbf{x}_i$ at timestep $i$ is an $N$-dimensional vector.

Textual data is defined as $\mathbf{X}_\text{Text} = \{s_1, s_2, \ldots, s_L\}$, where each symbol $s_i \in \mathcal{V}$ is drawn from a predefined vocabulary $\mathcal{V}$, and $L$ denotes the length of the text sequence. Specifically,
(1) \textbf{Textual time series}: This type represents numerical time series converted into discrete symbolic sequences, allowing LLMs to process time series in textual form~\cite{Survey_CMA}.
(2) \textbf{Natural language}: This type is not derived from time series conversion, including domain knowledge, task instructions, questions, and answers.

\begin{figure*}[!t]
\centering
\includegraphics[width=\linewidth]{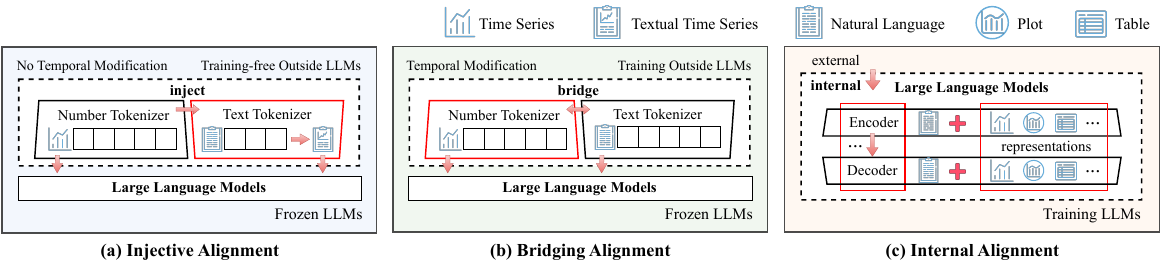}
\caption{Comparison of three alignment paradigms. Red boxes denote focused components, and icons indicate time series representations.}
\label{fig:alignment}
\vspace{-0.8em}
\end{figure*}

\subsection{Time Series Analysis with LLMs}
These tasks follow a predefined processing with fixed objectives, where LLMs serve as modeling and reasoning components for TSA tasks. In addition to generating corresponding explanations, the objectives and outputs of representative analysis tasks are defined as follows.
(1) \textbf{Forecasting:} Given historical observations, LLMs predict future numerical values over a predefined horizon. The output is a numerical sequence with temporal order~\cite{TSFLLMs}.
(2) \textbf{Anomaly detection:} LLMs identify deviations from normal dynamics. The outputs are binary labels (normal and abnormal) and anomaly scores~\cite{AnomalyLLM}.
(3) \textbf{Classification:} LLMs assign each time series to one category from a predefined label set. The output consists of one or more predefined class labels~\cite{InstructTime}.
(4) \textbf{Imputation:} LLMs infer missing or corrupted values using observed context. The output is a completed numerical sequence with missing entries filled~\cite{Survey_Imputation}.
Formally, given a time series $\mathbf{X}$ and contextual information $\mathbf{C}$ (e.g., natural language, plots), TSA aims to produce the task-specific output $Y$:

\[
f_p(\mathbf{X}, \mathbf{C}) \rightarrow Y
\]

Where $f_p(\cdot)$ refers to a method that uses LLMs with predefined task prefixes, and $p$ denotes the corresponding predefined task prompts, e.g., “predict the next 48 timesteps”.

\subsection{Time Series Question Answering}
In contrast to predefined analysis tasks, TSQA formulates time series understanding as a unified question answering paradigm. This paradigm enables flexible exploration of time series while emphasizing:
(1) \textbf{Deep thinking}: analyzing complex temporal semantics and dependencies to derive clear insights~\cite{TimeSense}.
(2) \textbf{Action planning}: decomposing user queries into goal-oriented steps to generate optimal answers~\cite{Survey_AgenticTSLLM}.
(3) \textbf{Tool calling}: using external analytical tools or models to perform precise numerical computations, statistical tests, and other operations within the reasoning process~\cite{STReasoner}.
Under this paradigm, representative TSQA tasks include:
(1) \textbf{In-depth interpretation}: explaining user questions through joint reasoning over temporal patterns, multivariate dependencies, and contextual information.
(2) \textbf{Time series captioning}: producing natural language descriptions that summarize temporal patterns, structure, and events~\cite{TSLM}.

Formally, given a time series $\mathbf{X}$, additional contextual information $\mathbf{C}$, and a natural language question $Q$, the goal of TSQA is to produce the correct answer $A$:

\[
f(\mathbf{X}, \mathbf{C}, Q) \rightarrow A
\]

Where $f(\cdot)$ refers to a method that uses LLMs with $Q$, and $A$ is a natural language answer that leverages temporal semantics from $\mathbf{X}$ and relevant $\mathbf{C}$~\cite{Time-MQA}.

\section{Taxonomy}
\label{sec:taxonomy}
In this section, we propose a taxonomy with clear boundaries, formal definitions, optimization strategies, and categorize existing literature into three alignment paradigms.

\subsection{Overview}
The three alignment paradigms are Injective Alignment, Bridging Alignment, and Internal Alignment. As illustrated in Figure~\ref{fig:alignment}, the processes and emphases of each alignment paradigm are as follows:
(a) Injective Alignment focuses on textual representations by injecting numerical time series into textual prompts. 
(b) Bridging Alignment focuses on numerical representations and builds connections between numerical time series and textual prompts.
(c) Internal Alignment typically focuses on internal component modifications and multiple time series representations, including numerical values, visual plots, and structured tables. These paradigms cover the existing literature with clear boundaries, and each work can be uniquely assigned to a single category. 

\begin{figure}[!h]
\centering
\includegraphics[width=\linewidth]{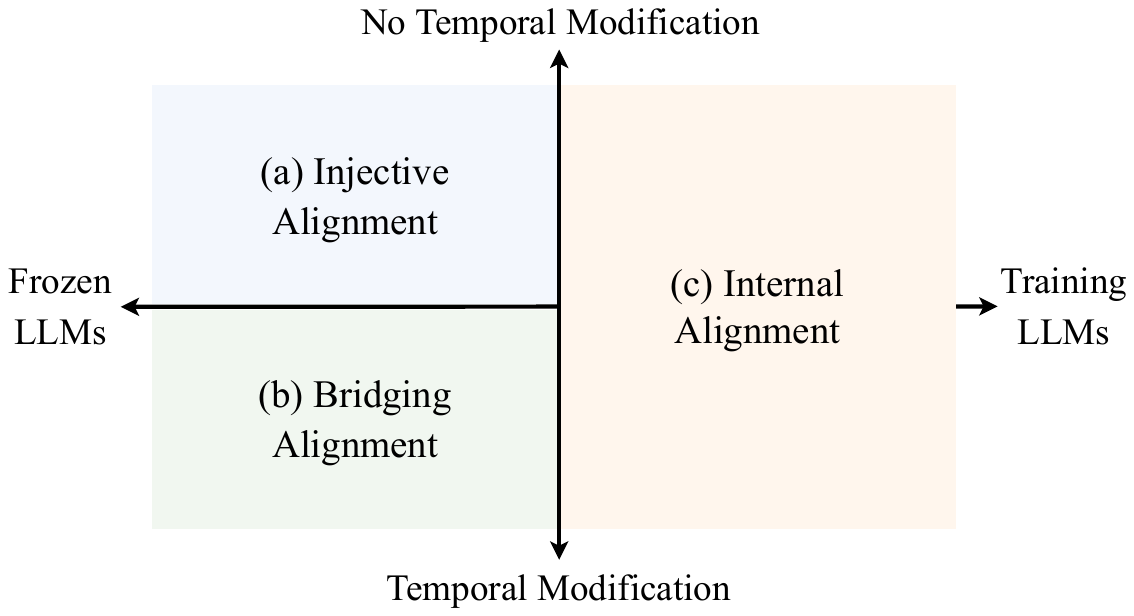}
\caption{Boundaries of three proposed alignment paradigms.}
\label{fig:boundary}
\vspace{-0.6em}
\end{figure}

As shown in Figure~\ref{fig:boundary}, the horizontal axis indicates whether LLM parameters are trained, and the vertical axis indicates whether temporal modifications are required. Temporal modification refers to adapting time series for LLMs, including both modifications outside the LLM and adjustments to the internal LLM architecture. These two dimensions determine the three alignment paradigms.
(a) Injective Alignment involves no temporal modification and adopts frozen LLMs. This design preserves the original LLM parameters.
(b) Bridging Alignment introduces temporal modification while still employing frozen LLMs. This design enables joint processing of time series and textual inputs while preserving all parameters of the original LLM.
(c) Internal Alignment combines temporal modification with training LLMs through parameter updating to provide native support for time series. Next, we detail the three paradigms along these boundaries.

\subsection{Injective Alignment}
This paradigm injects information into LLMs by directly encoding $\mathbf{X}_\text{Number}$ into token embeddings or appending textual time series $\mathbf{X}_\text{Text}$ to prompts. This process is defined as $\mathbf{Y}=f_{\theta}(h_{\gamma}(\mathbf{X}_\text{Number}), \mathbf{X}_\text{Text})$, where $h_{\gamma}(\cdot)$ is an injective method, and $f_{\theta}(\cdot)$ refers to frozen LLMs. Beyond original \textbf{Direct} input strategies, two optimization strategies for $h_{\gamma}(\cdot)$ are In-Context Learning (ICL) and Chain of Thought (CoT).

\textbf{ICL.}
This strategy guides LLMs with a few task demonstrations to enhance contextual understanding and reasoning. For instance,
LLMAD~\cite{LLMAD} further improves ICL through diversified demonstrations and robust retrieval, finding that a small mix of positive and negative examples achieves an effective trade-off, as additional demonstrations offer limited gains.
LVICL~\cite{LVICL} compresses multiple time-series examples into a permutation-invariant context vector and injects it into each LLM layer to elicit in-context learning for improved forecasting while keeping the LLM frozen.
\textbf{Benefits}: Effective in data-scarce scenarios and allows low-cost dynamic context switching.
\textbf{Drawbacks}: Performance is sensitive to noisy or low-quality examples.

\textbf{CoT.}
This strategy activates LLMs for deep thinking by decomposing complex tasks into ordered intermediate steps along explicit logical paths. For example, 
LSTPrompt~\cite{LSTPrompt} divides zero-shot forecasting of TSA into two stages: short-term reasoning for dynamic trends and long-term reasoning for periodic patterns, with periodic revisits to assumptions to enhance temporal adaptivity. 
TimerBed~\cite{TimerBed} applies a three-phase reasoning path for TSQA: it first identifies relevant temporal segments; then extracts numerical clues from these segments; it finally synthesizes explainable answers.
\textbf{Benefits}: Converts implicit logic into explicit steps, enhancing interpretability and reliability via stepwise self-correction.
\textbf{Drawbacks}: Requires predefined steps and incurs additional computational cost due to extra CoT. Practical instructions for Injective Alignment are as follows:
\begin{mybox}
This paradigm is \textbf{flexible} to operate outside LLMs. CoT and ICL are easy to implement across domains, making this paradigm suitable for domain-specific applications. Since its effectiveness depends on prompt quality, practitioners should iteratively refine the design of prompts using domain expertise and LLM feedback.
\end{mybox}

\subsection{Bridging Alignment}
This paradigm maps numerical time series and textual prompts, defined as $\mathbf{Y}=f_{\theta}(g_{\phi}(\mathbf{X_\text{Number}}, \mathbf{X_\text{Text}}))$, where $g_{\phi}(\cdot)$ is a bridging method and $f_{\theta}(\cdot)$ refers to frozen LLMs. Two efficient optimization strategies for $g_{\phi}(\cdot)$ are as follows.

\textbf{Similarity Retrieval (SR).} 
This strategy aligns numerical and textual representations via similarity. Adapters serve as a bridging space, projecting time series into an embedding space compatible with LLMs. Specifically, $g_{\phi}$ maps time series as $g_{\phi}(\mathbf{X_\text{Number}})\to\mathbf{X}_{\text{Text}}$, with $\phi$ trained using a loss function $\mathcal{L}$. For example, 
Time-LLM~\cite{Time-LLM} introduces a reprogramming space that converts time series into text prototypes, bridging numeric encodings with textual embeddings.
Time-CMA~\cite{TimeCMA} encodes time series and text into a shared space using separate encoders, aligns embeddings via contrastive loss, and retrieves time series whose embedding best matches the query.
\textbf{Benefits:} Enhances performance via flexible adapters without altering LLM parameters.
\textbf{Drawbacks:} Performance depends on adapter quality and incurs computational cost during retrieval.

\textbf{Low-Rank Adaptation (LoRA).} 
This strategy is a parameter-efficient technique that bridges low-rank updates into pre-trained LLMs, enabling task-specific tuning without altering the model parameters. For instance, 
HiTime~\cite{HiTime} applies LoRA during generative instruction fine-tuning for time series classification, updating a small subset of weights to enable generative reasoning from numeric features. 
AutoTimes~\cite{AutoTimes} integrates LoRA into its autoregressive forecasting framework, fine-tuning embedding and projection layers while keeping the LLM parameters frozen. 
\textbf{Benefits:} Parameter-efficient and task-specific adaptation.
\textbf{Drawbacks:} Limited performance when the low-rank space is insufficient for complex time-series patterns. Practical instructions for Bridging Alignment are as follows:

\begin{mybox}
This paradigm is \textbf{economical} to operate outside LLMs. Similarity Retrieval and LoRA leverage aligned semantic information, making it effective for capturing time-series characteristics. Since its effectiveness depends on the quality of adapter modules, practitioners should carefully assess whether the improvements are meaningful.
\end{mybox}

\renewcommand{\tiny}{\fontsize{8}{10}\selectfont}

\tikzstyle{my-box}=[
    rectangle,
    draw=hidden-draw,
    rounded corners,
    text opacity=1,
    minimum height=1.5em,
    minimum width=5em,
    inner sep=2pt,
    align=center,
    fill opacity=.5,
    line width=0.8pt,
]
\tikzstyle{leaf}=[my-box, minimum height=1.5em,
    fill=hidden-pink!80, text=black, align=left, font=\tiny,
    inner xsep=2pt,
    inner ysep=4pt,
    line width=0.8pt,
]
\begin{figure*}[!h]
    \centering
    \resizebox{1\textwidth}{!}{
        \begin{forest}
            forked edges,
            for tree={
                fill=level0!80,
                grow=east,
                reversed=true,
                anchor=base west,
                parent anchor=east,
                child anchor=west,
                base=left,
                font=\tiny,
                rectangle,
                draw=hidden-draw,
                rounded corners,
                align=center,
                minimum width=4em,
                edge+={darkgray, line width=1pt},
                s sep=3pt,
                inner xsep=2pt,
                inner ysep=3pt,
                line width=0.8pt,
                ver/.style={rotate=90, child anchor=north, parent anchor=south, anchor=center},
            },
            where level=1{text width=3.5em, font=\tiny, fill=level1!50}{},
            where level=2{text width=2.2em, font=\tiny, fill=level2!50, align=center, minimum width=2.2em}{},
            where level=3{text width=3.2em, font=\tiny, fill=level3!50, align=center, minimum width=3.2em}{},
            where level=4{text width=2.5em, font=\tiny, fill=level4!50, align=left}{}, 
            [Alignment\\Paradigms
                [
                    Injective\\Alignment
                    [~TSA
                        [
                            Financial
                            [
                                {\textbf{Direct:}~Instruct-FinGPT~\cite{Instruct-FinGPT}.}, text width=23em
                            ]
                        ]
                        [
                            ~General
                            [
                                {\textbf{Direct:}~TimerBed~\cite{TimerBed}.}, text width=23em
                            ]
                            [
                                {\textbf{ICL:}~LVICL~\cite{LVICL}.}, text width=23em
                            ]
                        ]
                    ]
                    [TSQA
                        [
                            ~Medical 
                            [
                                {\textbf{Direct:}~MedTsLLM~\cite{MedTsLLM}.}, text width=23em
                            ]
                        ]   
                        [
                            ~General 
                            [
                                {\textbf{Direct:}~Argos~\cite{Argos}, TableTime~\cite{TableTime},\\ TSFLLMs~\cite{TSFLLMs}, PromptCast~\cite{PromptCast}.}, text width=23em
                            ]
                            [
                                {\textbf{CoT:}~AnomLLM~\cite{AnomLLM}, TEMPO~\cite{TEMPO},\\
                                LSTPrompt~\cite{LSTPrompt}, 
                                LLMTime~\cite{LLMTime}.}, text width=23em
                            ]
                            [
                                {\textbf{ICL:}~LLMAD~\cite{LLMAD}.}, text width=23em
                            ]
                        ]
                    ]
                ]
                [
                    Bridging\\Alignment
                    [
                        ~TSA
                        [
                            ~Medical
                            [
                                {\textbf{SR:} ConMIL~\cite{ConMIL}, DeWave~\cite{DeWave},\\ ECG-LLM~\cite{ECG-LLM}.}, text width=23em
                            ]
                        ]
                        [
                            Financial
                            [
                                {\textbf{SR:} SCRL-LG~\cite{SCRL-LG}.}, text width=23em
                            ]
                        ]
                        [
                            ~General
                            [
                                {\textbf{SR:}~Time-VLM~\cite{Time-VLM}, 
                                MedualTime~\cite{MedualTime},\\ 
                                TimeCAP~\cite{TimeCAP}, TSLM~\cite{TSLM},\\ 
                                LLM-TPF~\cite{LLM-TPF}, TimeCMA~\cite{TimeCMA},\\
                                TimeXL~\cite{TimeXL}, TimeKD~\cite{TimeKD},\\
                                Time-LLM~\cite{Time-LLM}, S$^2$IP-LLM~\cite{S2IP-LLM},\\
                                LeRet~\cite{LeRet}, UniTime~\cite{UniTime},\\ 
                                Time-FFM~\cite{Time-FFM}, AnomalyLLM~\cite{AnomalyLLM}.}, text width=23em
                            ]
                            [
                                {\textbf{LoRA:}~CALF~\cite{CALF}, Time-LlaMA~\cite{Time-LlaMA},\\
                                LLM4TS~\cite{LLM4TS}, HiTime~\cite{HiTime},\\
                                AutoTimes~\cite{AutoTimes}.}, text width=23em
                            ]
                        ]
                    ]
                    [
                        TSQA
                        [
                            ~~~IoT
                            [
                                {\textbf{LoRA:}~SensorChat~\cite{SensorChat}, SensorLLM~\cite{SensorLLM}.}, text width=23em
                            ]
                        ] 
                    ]
                ]
                [
                    Internal\\Alignment
                    [
                        ~TSA
                        [
                            ~Medical 
                            [
                                {\textbf{RT:}~ECG-Chat~\cite{ECG-Chat}.}, text width=23em
                            ]
                        ]
                        [
                            ~~~IoT
                            [
                                {\textbf{RT:}~GenG~\cite{GenG}.}, text width=23em
                            ]
                        ]
                        [
                            ~General
                            [
                                {\textbf{RT:}~ChatTime~\cite{ChatTime}.}, text width=23em
                            ]
                            [
                                {\textbf{RL:}~Alpha-R1~\cite{Alpha-R1}.}, text width=23em
                            ]
                        ]
                    ]
                    [
                        TSQA
                        [
                            Financial
                            [
                                {\textbf{RT:}~FinTral~\cite{FinTral}, BloombergGPT~\cite{BloombergGPT}.}, text width=23em
                            ]
                        ]
                        [
                            ~~~IoT
                            [
                                {\textbf{RT:}~CityGPT~\cite{CityGPT}.}, text width=23em
                            ]
                        ]
                        [
                            ~General
                            [
                                {\textbf{RT:}~ChatTS~\cite{ChatTS}, Time-MQA~\cite{Time-MQA},\\  InstructTime~\cite{InstructTime}.}, text width=23em
                            ]
                            [
                                {\textbf{RL:}~STReasoner~\cite{STReasoner}, TimeSense~\cite{TimeSense}.}, text width=23em
                            ]
                        ]
                    ]
                ]
            ]
        \end{forest}
    }
    \caption{A comprehensive taxonomy of relevant literature, organized by three alignment paradigms, task types, domains, and optimization strategies. Notably, the category with the most literature is the Bridging Alignment, TSA task, General domain, and SR optimization strategy, where the General domain denotes literature without a specific application focus and evaluated across multiple domains.}
    \label{fig:literature}
\end{figure*}

\subsection{Internal Alignment}
This paradigm modifies internal components and parameters, defined as ${\mathbf{Y}={f}'_{\theta}(d_{\psi1}(\mathbf{X_\text{Number}}), d_{\psi2}(\mathbf{X_\text{Text}}),\dots)}$, where ${\{d_{\psi1}(\cdot),d_{\psi2}(\cdot),\ldots\}}$ are internal methods that enhance $\mathbf{X}{}_{\text {Number}}$ into $\mathbf{X}{}'_{\text {Number}}$ and $\mathbf{X}{}_{\text{Text}}$ into $\mathbf{X}{}'_{\text{Text}}$, or introduces additional representations. ${f}'_{\theta}(\cdot)$ denotes training LLMs. Two representative optimization strategies for ${f}'_{\theta}(\cdot)$ are as follows.
 
\textbf{Re-Training (RT).}
This strategy aligns time series with LLMs by temporal modification and updating parameters. For example, 
ChatTime~\cite{ChatTime} treats time series as a foreign language. It normalizes and discretizes continuous signals into symbolic tokens, extends the LLM vocabulary, and performs continual pre-training and instruction tuning to unify temporal and textual modalities.
ChatTS~\cite{ChatTS} trains a native time series LLM for TSQA. It synthesizes high-quality temporal descriptions using rule-based generators and attribute selectors, followed by two-stage fine-tuning to align numerical evolution with language reasoning.
\textbf{Benefits}: Enables deep internal alignment between temporal patterns and linguistic representations, and improves multi-task adaptability via parameter updates.
\textbf{Drawbacks}: Requires substantial training resources and may introduce additional complexity from discretization or synthetic data.

\textbf{Reinforcement Learning (RL).}
This strategy aligns LLM behavior with temporal objectives by optimizing decision policies using reward signals derived from sequential outcomes. For instance, 
TimeSense~\cite{TimeSense} treats temporal reasoning as a self-supervised reconstruction task. It uses time-aware objectives to learn temporal order, duration, and dependencies, which implicitly ensure temporal consistency via reinforcement learning.
Alpha-R1~\cite{Alpha-R1} models factor selection as a multi-step decision trajectory and directly optimizes LLM outputs via reinforcement learning. This formulation grounds language generation in long-horizon temporal feedback and aligns reasoning steps with profit-driven sequential outcomes.
\textbf{Benefits}: Directly optimizes long-horizon temporal objectives, adapts to dynamic environments, and provides decision interpretability through reward semantics.
\textbf{Drawbacks}: Training stability depends on reward design, optimization cost is high, and performance is sensitive to data noise and quality. Practical instructions for Internal Alignment are as follows:
\begin{mybox}
This paradigm is \textbf{generalizable} for internal LLM operation. Re-Training and Reinforcement Learning provide general-purpose improvement for time series understanding and reasoning. Since its effectiveness depends on the dataset and retraining quality, practitioners should focus on data synthesis methods and evaluation metrics.
\end{mybox}

\begin{figure}[!h]
\centering
\includegraphics[width=1\linewidth]{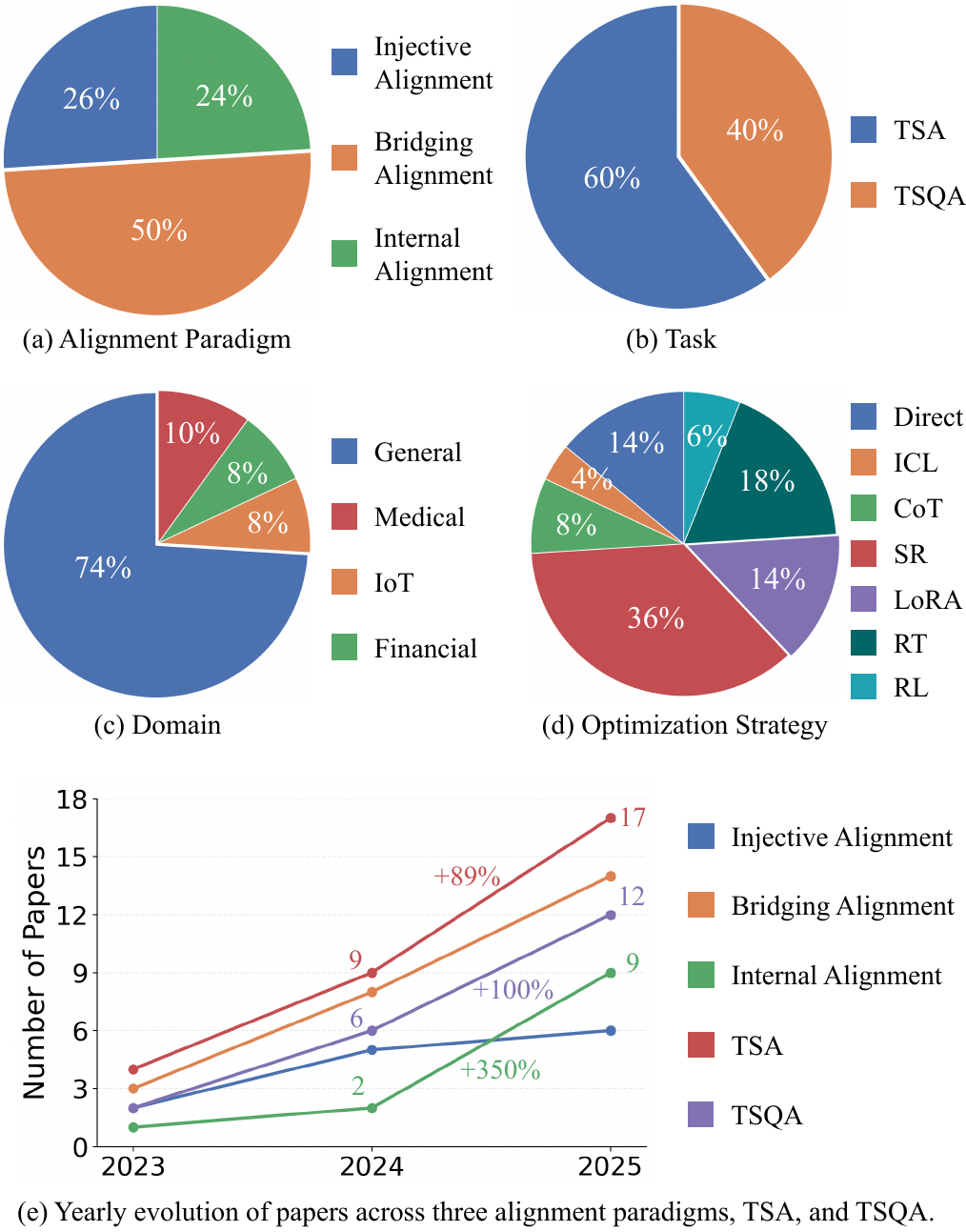}
\caption{Taxonomy distribution and evolution across dimensions.}
\label{fig:pie}
\vspace{-1em}
\end{figure}

We categorize 50 relevant literature into three groups from TSA to TSQA, as shown in Figure~\ref{fig:literature}. For each work, a comprehensive taxonomy is summarized across four dimensions of Alignment Paradigms, tasks, domains, and optimization strategies.
Specifically, we analyze the distribution of Alignment Paradigms in Figure~\ref{fig:pie}(a), including Injective Alignment 26\%, Bridging Alignment 50\%, and Internal Alignment 24\%. We further examine the task distribution in Figure~\ref{fig:pie}(b), which consists of TSA 60\% and TSQA 40\%. As shown in Figure~\ref{fig:pie}(c), we identify the application domains covered in the literature, including Medical 10\%, Finance 8\%, IoT 8\%, and General 74\%. In addition, we focus on the optimization strategies summarized in Figure~\ref{fig:pie}(d), where Similarity Retrieval (SR) accounts for the largest proportion at 36\%. We present the yearly evolution trends in Figure~\ref{fig:pie}(e), with TSQA surpassing TSA in growth rate from 2024 to 2025, and Internal Alignment exhibiting the most significant increase.
Next, we analyze representative datasets across diverse domains, highlighting how question answering design grounds time series characteristics and drives their evolution.

\begin{table*}[ht]
\vspace{-0.1em}
\centering
\begin{tabular}{@{}l|c|c|c|c@{}}
\toprule
\rowcolor[HTML]{EFEFEF} 
\textbf{Dataset} & \textbf{Domain} & \textbf{Task} & \textbf{Size} & \textbf{Characteristic} \\ \midrule
ECG-QA\cite{Dataset-ECGQA} 
& Medical
& TSQA
& 414,348 pairs.
& 3 types, 12 variates,
2 modalities.\\ \midrule

FinTexTS~\cite{FinTexTS}
& Financial
& TSA, TSQA
& 100M pairs.
& 4 variates, 2 modalities.\\ \midrule

SensorQA~\cite{SensorQA} 
& IoT
& TSQA
& 5,648 pairs.
& 2 types, multivariate,
5 modalities.\\ \midrule

Time-MQA~\cite{Time-MQA}
& \multirow{4}{*}{General}
& TSQA
& 192,843 pairs.
& 6 types, multivariate,
2 modalities.\\ \cmidrule(lr){1-1}\cmidrule(lr){3-5}

CiK~\cite{CiK} 
& 
& TSA
& 2,644 pairs.
& 5 types, univariate,
2 modalities.\\ \cmidrule(lr){1-1}\cmidrule(lr){3-5}

TSandLanguage~\cite{TSandLanguage} 
& 
& TSA, TSQA
& 230k pairs.
& 10 types, univariate,
2 modalities.\\
\bottomrule
\end{tabular}
\caption{Summary of datasets from TSA to TSQA, including Domain, Task, Size, and Characteristic (patterns, variables, and modalities).}
\label{tab:dataset}
\vspace{-0.5em}
\end{table*}

\section{Dataset Analysis}
\label{sec:dataset}
In this section, we introduce representative datasets organized by application domains, as shown in Table~\ref{tab:dataset}. Additional datasets are provided in our GitHub repository\footnote{\url{https://github.com/Leeway-95/TSA-TSQA-with-LLMs}}.

\subsection{Overview}
We analyze representative datasets from two perspectives: application domains and data characteristics, which respectively determine their semantics and structures. From the domain perspective, existing datasets vary in semantic dependence and regulatory constraints, but they generally require strong interpretability, contextual understanding, and integration of domain knowledge. From the characteristic perspective, existing datasets differ in temporal patterns and spatial multivariate structures. Moreover, time series can be represented in multiple forms, including numerical sequences, textual descriptions, and visual plots, each introducing different trade-offs among fidelity, interpretability, and compatibility.

\subsection{Domain}
Time series data span a wide range of domains, covering representative medical, financial, and IoT domains.

\textbf{Medical data.}
Medical time series include EEG, ECG, and related signals, which often exhibit strong seasonality and clear outliers. Semantic interpretation relies heavily on domain knowledge, as specific waveforms correspond to physiological states~\cite{DeWave}. Medical data are typically collected by professional equipment, but they are subject to strict privacy regulations and require transparent reasoning to support clinical decisions~\cite{MedTsLLM}.

\textbf{Financial data.}
Financial time series are high-frequency, such as stock prices and trading volumes. They contain pronounced outliers and short-term volatility, while also exhibiting recurring trends relevant to risk assessment and market forecasting~\cite{BloombergGPT}. Financial data usually rely on external knowledge and operate under stringent market regulations. Interpreting financial time series is often more valuable than predicting exact values~\cite{SCRL-LG}.

\textbf{IoT data.}
IoT time series consist of multi-source sensor data that typically show seasonality and infrequent changes due to predefined operating procedures. IoT data are high-dimensional and heterogeneous, integrating multiple sensors. Data quality is influenced by sensor type, sampling rate, and hardware-induced noise~\cite{GenG}. Effective TSQA in this domain requires contextual fusion to ensure comprehensive and consistent interpretation~\cite{SensorChat}.

\subsection{Characteristic}
Temporal patterns exhibit varying properties over time, including:
(1) \textbf{Outliers}, representing deviant signals with anomalous semantics~\cite{AnomLLM}.  
(2) \textbf{Trend}, capturing long-term directional changes over time~\cite{TimeXL}.  
(3) \textbf{Seasonality}, corresponding to recurring semantics at regular intervals~\cite{SensorChat}.  
(4) \textbf{Volatility}, characterized by irregular fluctuations that obscure meaningful patterns and often require denoising~\cite{S2IP-LLM}.  

Spatial multivariates introduce temporal complexity, including:
(1) \textbf{Independence}, treating variables separately to reduce complexity and avoid spurious correlations~\cite{TimeCMA}.
(2) \textbf{Dependence}, capturing inter-variable interactions by considering variables jointly~\cite{LLM-TPF}.
(3) \textbf{Correlation}, representing complex relational structures for coherent and context-aware reasoning~\cite{YunyaoCheng_1}.

Beyond temporal patterns and spatial multivariates, multimodal representations provide an additional characteristic.
(1) \textbf{Numerical representation} preserves raw time series values and offers advantages: it directly retains precise quantitative information and avoids additional transformation costs. However, numerical data also have limitations: time series are scarce compared to language, and raw values lack explicit semantic meaning~\cite{LLMTime}. Real-world signals often contain noise, missing values, and sparsity, requiring preprocessing.  
(2) \textbf{Textual representation} converts time series into natural language descriptions and offers advantages: text provides human-readable semantics that are intuitive and interpretable~\cite{TimeCAP}. This format naturally aligns with LLMs. Its main limitation lies in the alignment between the discrete nature of text and the continuous properties of time series signals~\cite{ChatTime}. 
(3) \textbf{Visual representation} draws time series as plots, such as line charts, and offers advantages: visual representations explicitly highlight trends, seasonality, and outliers. Visual plots can be aligned with LLMs via zero-shot visual understanding and can supplement or replace numerical or textual representations~\cite{Time-VLM}. Nevertheless, visual plots introduce challenges: low resolution may cause information loss, and fine-grained temporal values are often difficult to recover precisely, limiting quantitative accuracy~\cite{Survey_MTSA}. These characteristics jointly form the complexity of temporal semantics and modeling, requiring LLMs to capture temporal dynamics, inter-variable dependencies, and multimodal representations for temporal understanding and reasoning.

\section{Challenge and Future Direction}
\label{sec:future}
We further discuss the challenges and future directions under the three alignment paradigms proposed in this work.

\subsection{Data}
The evolution of datasets toward TSQA exposes several structural limitations.
(1) \textbf{Lack of reasoning supervision} is a primary bottleneck. Most existing datasets emphasize short answers, while providing little supervision over intermediate reasoning steps or tool usage. Future datasets should explicitly incorporate reasoning traces and expected temporal behaviors to support reasoning~\cite{QuAnTS}.
(2) \textbf{Insufficient domain diversity} constrains generalized evaluation. Time series have domain-specific variation in sequence length, granularity, noise patterns, and temporal dependencies. Constructing large-scale, multi-domain TSQA datasets with temporal diversity is critical~\cite{MTBench}.

\subsection{Task}
TSA and TSQA are complementary rather than mutually exclusive. TSA provides effective task-specific solutions, whereas TSQA offers a unified question answering paradigm based on temporal reasoning. This evolution from TSA to TSQA also introduces fundamental challenges.
(1) \textbf{Tasks shift from explicit specification to implicit extraction}.
TSA assumes clearly defined tasks with specified objectives. In contrast, TSQA identifies the time series problems from user inputs that typically contain underlying temporal semantics. This shift requires models to interpret user intent into appropriate tasks~\cite{HiTime}.
(2) \textbf{Lack of task tools for TSQA}.
Although TSQA extends TSA toward a unified question answering paradigm, it often lacks tools explicitly designed for diverse temporal questions, such as time series slicing~\cite{TimeSense}. Future TSQA should incorporate dedicated tools or invoke effective TSA models to enhance temporal reasoning~\cite{Time-MQA}.

\subsection{Method}
Several challenges arise across optimization strategies.
(1) \textbf{Limited adaptivity}. Current methods struggle with complexity and uncertainty. Future approaches should dynamically support long-horizon, multi-scale, and irregular time series under temporal reasoning~\cite{Argos}. Incorporating multi-agent mechanisms can further enable planning, tool use, and self-reflection, turning alignment into an adaptive process~\cite{Survey_AgenticTSLLM}. (2) \textbf{Numerical hallucination}. Current approaches struggle with numerical hallucination in TSA and TSQA. Discrete tokenization limits sensitivity to continuous values, producing plausible but incorrect reasoning. Future work should investigate how alignment paradigms mitigate this issue~\cite{TimeART}.

\section{Conclusion}
In this survey, we categorize existing literature into three paradigms, summarize optimization strategies along the evolution from TSA to TSQA, and provide practical guidance for flexible, economical, and generalizable selection of alignment paradigms. We further analyze representative datasets, identify challenges, and highlight future research directions.

\section*{Acknowledgments} This work is supported by the National Natural Science Foundation of China (Nos. 62172423 and 62402414) and the Guangdong Basic and Applied Basic Research Foundation (No. 2025A1515011994).

\appendix

\bibliographystyle{named}
\bibliography{ijcai26}

\end{document}